\documentclass[conference]{IEEEtran}
\usepackage{times}
\usepackage{latexsym}
\usepackage{booktabs}
\usepackage{amsmath}

\usepackage{color, graphicx, float, subfig}
\usepackage[dvipsnames]{xcolor}

\date{}

\begin{document}

\title{Event Outcome Prediction using Sentiment Analysis and Crowd Wisdom in Microblog Feeds}


\author{\IEEEauthorblockN{Rahul Radhakrishnan Iyer}
\IEEEauthorblockA{
	    Carnegie Mellon University\\
	    5000 Forbes Avenue\\
	    Pittsburgh, PA 15213, USA\\
	    rahuli@andrew.cmu.edu}
        \and
        \IEEEauthorblockN{Ronghuo Zheng}
	  	\IEEEauthorblockA{
	    Carnegie Mellon University\\
	    5000 Forbes Avenue\\
	    Pittsburgh, PA 15213, USA\\
	    ronghuoz@andrew.cmu.edu}
	    \and
       \IEEEauthorblockN{Yuezhang Li}
		\IEEEauthorblockA{
	    Carnegie Mellon University\\
	    5000 Forbes Avenue\\
	    Pittsburgh, PA 15213, USA\\
	    yuezhanl@andrew.cmu.edu}
	    \and
	  	\IEEEauthorblockN{Katia Sycara}
	  	\IEEEauthorblockA{
	    Carnegie Mellon University\\
	    5000 Forbes Avenue\\
	    Pittsburgh, PA 15213, USA\\
	    katia@cs.cmu.edu}}
	    
\maketitle
\begin{abstract}
 Sentiment Analysis of microblog feeds has attracted considerable interest in recent times. Most of the current work focuses on tweet sentiment classification. But not much work has been done to explore how reliable the opinions of the mass (crowd wisdom) in social network microblogs such as twitter are in predicting outcomes of certain events such as election debates. In this work, we investigate whether crowd wisdom is useful in predicting such outcomes and whether their opinions are influenced by the experts in the field. We work in the domain of multi-label classification to perform sentiment classification of tweets and obtain the opinion of the crowd. This learnt sentiment is then used to predict outcomes of events such as: US Presidential Debate winners, Grammy Award winners, Super Bowl Winners. We find that in most of the cases, the wisdom of the crowd does indeed match with that of the experts, and in cases where they don't (particularly in the case of debates), we see that the crowd's opinion is actually influenced by that of the experts.
\end{abstract}

\section{Introduction}
\label{sec:intro}
Over the past few years, microblogs have become one of the most popular online social networks. Microblogging websites have evolved to become a source of varied kinds of information. This is due to the nature of microblogs: people post real-time messages about their opinions and express sentiment on a variety of topics, discuss current issues, complain, etc. Twitter is one such popular microblogging service where users create status messages (called ``tweets"). With over $400$ million tweets per day on Twitter, microblog users generate large amount of data, which cover very rich topics ranging from politics, sports to celebrity gossip. Because the user generated content on microblogs covers rich topics and expresses sentiment/opinions of the mass, mining and analyzing this information can prove to be very beneficial both to the industrial and the academic community. Tweet classification has attracted considerable attention because it has become very important to analyze peoples' sentiments and opinions over social networks.

Most of the current work on analysis of tweets is focused on sentiment analysis ~\cite{go2009twitter,bermingham2010classifying,pak2010twitter}. Not much has been done on predicting outcomes of events based on the tweet sentiments, for example, predicting winners of presidential debates based on the tweets by analyzing the users' sentiments. This is possible intuitively because the sentiment of the users in their tweets towards the candidates is proportional to the performance of the candidates in the debate.

In this paper, we analyze three such events: 1) US Presidential Debates 2015-16, 2) Grammy Awards 2013, and 3) Super Bowl 2013. The main focus is on the analysis of the \textbf{presidential debates}. For the Grammys and the Super Bowl, we just perform sentiment analysis and try to predict the outcomes in the process. For the debates, in addition to the analysis done for the Grammys and Super Bowl, we also perform a trend analysis. Our analysis of the tweets for the debates is $3$-fold as shown below.

\textbf{Sentiment:} Perform a sentiment analysis on the debates. This involves: building a machine learning model which learns the sentiment-candidate pair (candidate is the one to whom the tweet is being directed) from the training data and then using this model to predict the sentiment-candidate pairs of new tweets.

\textbf{Predicting Outcome:} Here, after predicting the sentiment-candidate pairs on the new data, we predict the winner of the debates based on the sentiments of the users.

\textbf{Trends:} Here, we analyze certain trends of the debates like the change in sentiments of the users towards the candidates over time (hours, days, months) and how the opinion of experts such as \textit{Washington Post} affect the sentiments of the users.

For the sentiment analysis, we look at our problem in a multi-label setting, our two labels being sentiment polarity and the candidate/category in consideration. We test both single-label classifiers and multi-label ones on the problem and as intuition suggests, the multi-label classifier RaKel performs better. A combination of document-embedding features \cite{le2014distributed} and topic features (essentially the document-topic probabilities) \cite{blei2003latent} is shown to give the best results. These features make sense intuitively because the document-embedding features take context of the text into account, which is important for sentiment polarity classification, and topic features take into account the topic of the tweet (who/what is it about). 

The prediction of outcomes of debates is very interesting in our case. Most of the results seem to match with the views of some experts such as the political pundits of \textit{the Washington Post}. This implies that certain rules that were used to score the candidates in the debates by said-experts were in fact reflected by reading peoples' sentiments expressed over social media. This opens up a wide variety of learning possibilities from users' sentiments on social media, which is sometimes referred to as \emph{the wisdom of crowd}.

We do find out that the public sentiments are not always coincident with the views of the experts. In this case, it is interesting to check whether the views of the experts can affect the public, for example, by spreading through the social media microblogs such as Twitter. Hence, we also conduct experiments to compare the public sentiment before and after the experts' views become public and thus notice the impact of the experts' views on the public sentiment. In our analysis of the debates, we observe that in certain debates, such as the $5^{th}$ Republican Debate, held on December 15, 2015, the opinions of the users vary from the experts. But we see the effect of the experts on the sentiment of the users  by looking at their opinions of the same candidates the next day.

Our contributions are mainly: we want to see how predictive the sentiment/opinion of the users are in social media microblogs and how it compares to that of the experts. In essence, we find that the crowd wisdom in the microblog domain matches that of the experts in most cases. There are cases, however, where they don't match but we observe that the crowd's sentiments are actually affected by the experts. This can be seen in our analysis of the presidential debates.

The rest of the paper is organized as follows: in section \ref{sec:related}, we review some of the literature. In section \ref{sec:data}, we discuss the collection and preprocessing of the data. Section \ref{sec:tech_approach} details the approach taken, along with the features and the machine learning methods used. Section \ref{sec:results} discusses the results of the experiments conducted and lastly section \ref{sec:conclusions} ends with a conclusion on the results including certain limitations and scopes for improvement to work on in the future.

\section{Related Work}
\label{sec:related}
Sentiment analysis as a Natural Language Processing task has been handled at many levels of granularity. Specifically on the microblog front, some of the early results on sentiment analysis are by ~\cite{go2009twitter,bermingham2010classifying,pak2010twitter,barbosa2010robust,agarwal2011sentiment}. Go et al. ~\cite{go2009twitter} applied distant supervision to classify tweet sentiment by using emoticons as noisy labels. Kouloumpis et al. ~\cite{kouloumpis2011twitter} exploited hashtags in tweets to build training data. Chenhao Tan et al. ~\cite{tan2011user} determined user-level sentiments on particular topics with the help of the social network graph.

There has been some work in event detection and extraction in microblogs as well. In ~\cite{li2014major}, the authors describe a way to extract major life events of a user based on tweets that either congratulate/offer condolences. ~\cite{sayyadi2009event} build a key-word graph from the data and then detect communities in this graph (cluster) to find events. This works because words that describe similar events will form clusters. In ~\cite{reschke2014event}, the authors use distant supervision to extract events. There has also been some work on event retrieval in microblogs ~\cite{metzler2012structured}. In ~\cite{chierichetti2014event}, the authors detect time points in the twitter stream when an important event happens and then classify such events based on the sentiments they evoke using only non-textual features to do so. In \cite{wong2012watching}, the authors study how much of the opinion extracted from Online Social Networks (OSN) user data is reflective of the opinion of the larger population. Researchers have also mined Twitter dataset to analyze public reaction to various events: from election debate performance \cite{diakopoulos2010characterizing}, where the authors demonstrate visuals and metrics that can be used to detect sentiment pulse, anomalies in that pulse, and indications of controversial topics that can be used to inform the design of visual analytic systems for social media events, to movie box-office predictions on the release day \cite{asur2010predicting}. Mishne and Glance \cite{mishne2006predicting} correlate sentiments in blog posts with movie box-office scores. The correlations they observed for positive sentiments are fairly low and not sufficient to use for predictive purposes. Recently, several approaches involving machine learning and deep learning have also been used in the visual and language domains \cite{iyer2016content,li2016joint,iyer2018transparency,li2018object,gupta2016analysis,honke2018photorealistic,iyer2017detecting}.\\

\section{Data Set and Preprocessing}
\label{sec:data}
\subsection{Data Collection}
\label{subsec:data_coll}

Twitter is a social networking and microblogging service that allows users to post real-time messages, called tweets. Tweets are very short messages, a maximum of $140$ characters in length. Due to such a restriction in length, people tend to use a lot of acronyms, shorten words etc. In essence, the tweets are usually very noisy. There are several aspects to tweets such as: 1) Target: Users use the symbol ``@" in their tweets to refer to other users on the microblog. 2) Hashtag: Hashtags are used by users to mark topics. This is done to increase the visibility of the tweets. 

We conduct experiments on $3$ different datasets, as mentioned earlier: 1) US Presidential Debates, 2) Grammy Awards 2013, 3) Superbowl 2013. To construct our presidential debates dataset, we have used the Twitter Search API to collect the tweets. Since there was no publicly available dataset for this, we had to collect the data manually. The data was collected on $10$ different presidential debates: $7$ republican and $3$ democratic, from October $2015$ to March $2016$. Different hashtags like ``\#GOP, \#GOPDebate'' were used to filter out tweets specific to the debate. This is given in Table \ref{table:debate_data}. We extracted only english tweets for our dataset. We collected a total of $104961$ tweets were collected across all the debates. But there were some limitations with the API. Firstly, the server imposes a rate limit and discards tweets when the limit is reached. The second problem is that the API returns many duplicates. Thus, after removing the duplicates and irrelevant tweets, we were left with a total of $17304$ tweets. This includes the tweets only on the day of the debate. We also collected tweets on the days following the debate.

As for the other two datasets, we collected them from available-online repositories. There were a total of $2580062$ tweets for the Grammy Awards 2013, and a total of $2428391$ tweets for the Superbowl 2013. The statistics are given in Tables \ref{table:stats_grammy} and \ref{table:stats_superbowl}. The tweets for the grammy were before the ceremony and during. However, we only use the tweets before the ceremony (after the nominations were announced), to predict the winners. As for the superbowl, the tweets collected were during the game. But we can predict interesting things like \textit{Most Valuable Player} etc. from the tweets. The tweets for both of these datasets were annotated and thus did not require any human intervention. However, the tweets for the debates had to be annotated.

\begin{table}
\vspace{2ex}
\begin{tabular}{l | l | l}
\toprule
\textbf{Sl. No.} & \textbf{Date of Debate} & \textbf{Party}\\
\midrule
1 & October 28, 2015 & Republican \\
2 & November 10, 2015 & Republican \\
3 & December 15, 2015 & Republican \\
4 & January 14, 2016 & Republican \\
5 & January 17, 2016 & Democratic \\
6 & January 28, 2016 & Republican \\
7 & February 4, 2016 & Democratic \\
8 & February 25, 2016 & Republican\\
9 & March 9, 2016 & Democratic\\
10 & March 10, 2016 & Republican\\
\bottomrule
\end{tabular}
\centering
\caption{Debates chosen, listed in chronological order. A total of $10$ debates were considered out of which $7$ are Republican and $3$ are Democratic.}
\label{table:debate_data}
\end{table}

\begin{table}
\vspace{2ex}
\setlength\tabcolsep{4pt}
\begin{minipage}{0.48\textwidth}
\begin{tabular}{l | l }
\toprule
\textbf{Parameter} & \textbf{Value}\\
\midrule
Total Number of Tweets & 17304\\
Number of Debates & 10\\
Average Number of Tweets/Debate & 1730\\
Training Set Size & 10000\\
Testing Set Size & 7304\\
\bottomrule
\end{tabular}
\centering
\caption{Statistics of the Data Collected: Debates}
\label{table:stats_debates}

\end{minipage}%
\hfill
\begin{minipage}{0.48\textwidth}
\begin{tabular}{l | l }
\toprule
\textbf{Parameter} & \textbf{Value}\\
\midrule
Total Number of Tweets & 2580062\\
Training Set Size & 100000\\
Testing Set Size & 2480062\\
\bottomrule
\end{tabular}
\centering
\caption{Statistics of the Data Collected: Grammys 2013}
\label{table:stats_grammy}

\end{minipage}%
\hfill
\begin{minipage}{0.48\textwidth}
\begin{tabular}{l | l }
\toprule
\textbf{Parameter} & \textbf{Value}\\
\midrule
Total Number of Tweets & 2428391\\
Training Set Size & 100000\\
Testing Set Size & 2328391\\
\bottomrule
\end{tabular}
\centering
\caption{Statistics of the Data Collected: Superbowl 2013}
\label{table:stats_superbowl}
\end{minipage}
\end{table}



Since we are using a supervised approach in this paper, we have all the tweets (for debates) in the training set human-annotated. The tweets were already annotated for the Grammys and Super Bowl. Some statistics about our datasets are presented in Tables \ref{table:stats_debates}, \ref{table:stats_grammy} and \ref{table:stats_superbowl}. The annotations for the debate dataset comprised of $2$ labels for each tweet: 1) \textbf{Candidate: } This is the candidate of the debate to whom the tweet refers to, 2) \textbf{Sentiment: } This represents the sentiment of the tweet towards that candidate. This is either positive or negative.

The task then becomes a case of multi-label classification. The candidate labels are not so trivial to obtain, because there are tweets that do not directly contain any candidates' name. For example, the tweets, ``a business man for president??'' and ``a doctor might sure bring about a change in America!'' are about Donal Trump and Ben Carson respectively. Thus, it is meaningful to have a multi-label task.

The annotations for the other two datasets are similar, in that one of the labels was the sentiment and the other was category-dependent in the outcome-prediction task, as discussed in the sections below. For example, if we are trying to predict the "Album of the Year" winners for the Grammy dataset, the second label would be the nominees for that category (album of the year).

\subsection{Preprocessing}
\label{subsec:preprocess}
As noted earlier, tweets are generally noisy and thus require some preprocessing done before using them. Several filters were applied to the tweets such as: (1) \textbf{Usernames:} Since users often include usernames in their tweets to direct their message, we simplify it by replacing the usernames with the token ``USER''. For example, @michael will be replaced by USER. (2) \textbf{URLs:} In most of the tweets, users include links that add on to their text message. We convert/replace the link address to the token ``URL''. (3) \textbf{Repeated Letters:} Oftentimes, users use repeated letters in a word to emphasize their notion. For example, the word ``lol'' (which stands for ``laugh out loud'') is sometimes written as ``looooool'' to emphasize the degree of funnyness. We replace such repeated occurrences of letters (more than $2$), with just $3$ occurrences. We replace with $3$ occurrences and not $2$, so that we can distinguish the exaggerated usage from the regular ones. (4) \textbf{Multiple Sentiments:} Tweets which contain multiple sentiments are removed, such as "I hate Donald Trump, but I will vote for him". This is done so that there is no ambiguity. (5) \textbf{Retweets:} In Twitter, many times tweets of a person are copied and posted by another user. This is known as retweeting and they are commonly abbreviated with ``RT''. These are removed and only the original tweets are processed. (6) \textbf{Repeated Tweets:} The Twitter API sometimes returns a tweet multiple times. We remove such duplicates to avoid putting extra weight on any particular tweet.

\section{Methodology}
\label{sec:tech_approach}
\subsection{Procedure}
\label{subsec:procedure}
Our analysis of the debates is $3$-fold including sentiment analysis, outcome prediction, and trend analysis.

\textbf{Sentiment Analysis:} To perform a sentiment analysis on the debates, we first extract all the features described below from all the tweets in the training data. We then build the different machine learning models described below on these set of features. After that, we evaluate the output produced by the models on unseen test data. The models essentially predict candidate-sentiment pairs for each tweet.
		
\textbf{Outcome Prediction:} Predict the outcome of the debates. After obtaining the sentiments on the test data for each tweet, we can compute the net normalized sentiment for each presidential candidate in the debate. This is done by looking at the number of positive and negative sentiments for each candidate. We then normalize the sentiment scores of each candidate to be in the same scale ($0$-$1$). After that, we rank the candidates based on the sentiment scores and predict the top $k$ as the winners.
		
\textbf{Trend Analysis:} We also analyze some certain trends of the debates. Firstly, we look at the change in sentiments of the users towards the candidates over time (hours, days, months). This is done by computing the sentiment scores for each candidate in each of the debates and seeing how it varies over time, across debates. Secondly, we examine the effect of Washington Post on the views of the users. This is done by looking at the sentiments of the candidates (to predict winners) of a debate before and after the winners are announced by the experts in Washington Post. This way, we can see if Washington Post has had any effect on the sentiments of the users. Besides that, to study the behavior of the users, we also look at the correlation of the tweet volume with the number of viewers as well as the variation of tweet volume over time (hours, days, months) for debates.

As for the Grammys and the Super Bowl, we only perform the sentiment analysis and predict the outcomes.

\subsection{Machine Learning Models}
\label{subsec:ml_methods}
We compare $4$ different models for performing our task of sentiment classification. We then pick the best performing model for the task of outcome prediction. Here, we have two categories of algorithms: single-label and multi-label (We already discussed above why it is meaningful to have a multi-label task earlier), because one can represent $<$candidate, sentiment$>$ as a single class label or have candidate and sentiment as two separate labels. They are listed below:

\subsubsection{Single-label Classification}

\textbf{Naive Bayes}: We use a multinomial Naive Bayes model. A tweet $t$ is assigned a class $c^{*}$ such that
			\begin{align}
					c^{*} & = argmax_c P(c|t), \\
 					P(c|t) & = \frac{P(c) \times \sum_{i=1}^{m} P(f_i|c)}{P(t)},
			\end{align}
where there are $m$ features and $f_i$ represents the $i^{th}$ feature.

\textbf{Support Vector Machines}: Support Vector Machines (SVM) constructs a hyperplane or a set of hyperplanes in a high-dimensional space, which can then be used for classification. In our case, we use SVM with Sequential Minimal Optimization (SMO) \cite{platt1998sequential}, which is an algorithm for solving the quadratic programming (QP) problem that arises during the training of SVMs. 

\textbf{Elman Recurrent Neural Network}: Recurrent Neural Networks (RNNs) are gaining popularity and are being applied to a wide variety of problems. They are a class of artificial neural networks, where connections between units form a directed cycle. This creates an internal state of the network which allows it to exhibit dynamic temporal behavior. The Elman RNN was proposed by Jeff Elman in the year 1990 \cite{elman1990finding}. We use this in our task.

\subsubsection{Multi-label Classification}
\textbf{RAkEL} (RAndom k labELsets): RAkEL \cite{tsoumakas2011random} is a multi-label classification algorithm that uses labeled powerset (LP) transformation: it basically creates a single binary classifier for every label combination and then uses multiple LP classifiers, each trained on a random subset of the actual labels, for classification.

\subsection{Feature Space}
\label{subsec:features}
In order to classify the tweets, a set of features is extracted from each of the tweets, such as n-gram, part-of-speech etc. The details of these features are given below:
\begin{itemize}
	\item n-gram: This represents the frequency counts of n-grams, specifically that of unigrams and bigrams. 
	\item punctuation: The number of occurrences of punctuation symbols such as commas, exclamation marks etc.
	\item POS (part-of-speech): The frequency of each POS tagger is used as the feature.
	\item prior polarity scoring: Here, we obtain the prior polarity of the words \cite{agarwal2011sentiment} using the Dictionary of Affect in Language (DAL) \cite{whissell1989dictionary}. This dictionary (DAL) of about 8000 English words assigns a pleasantness score to each word on a scale of $1$-$3$. After normalizing, we can assign the words with polarity higher than $0.8$ as positive and less than $0.5$ as negative. If a word is not present in the dictionary, we lookup its synonyms in WordNet: if this word is there in the dictionary, we assign the original word its synonym's score.
	\item Twitter Specific features: 
	    \begin{itemize}
	        \item Number of hashtags ($\#$ symbol)
	        \item Number of mentioning users (\@ symbol)
	        \item Number of hyperlinks
	    \end{itemize}
	\item Document embedding features: Here, we use the approach proposed by Mikolov et al. \cite{le2014distributed} to embed an entire tweet into a vector of features
	\item Topic features: Here, LDA (Latent Dirichlet Allocation) \cite{blei2003latent} is used to extract topic-specific features for a tweet (document). This is basically the topic-document probability that is outputted by the model.
\end{itemize}

In the following experiments, we use $1$-$gram$, $2$-$gram$ and $(1+2)$-$gram$ to denote unigram, bigram and a combination of unigram and bigram features respectively. We also combine punctuation and the other features as miscellaneous features and use $MISC$ to denote this. We represent the document-embedding features by $DOC$ and topic-specific features by $TOPIC$.

\section{Data Analysis}
\label{sec:analysis}
In this section, we analyze the presidential debates data and show some trends. 

\begin{figure*}[!t]
\centering
\subfloat[Tweet Frequency across Debates]{\includegraphics[scale=0.25]{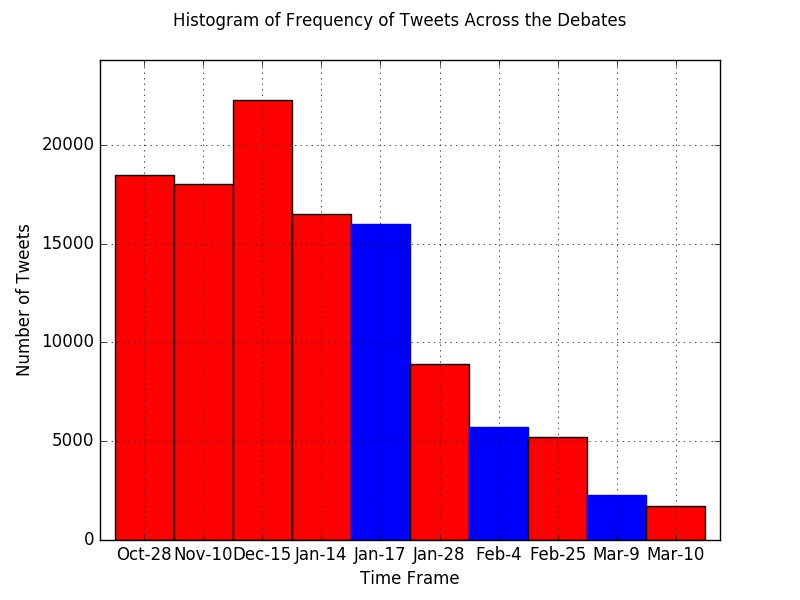}
\label{fig:tweet_debate}}
\hfil
\subfloat[Viewers across Debates]{\includegraphics[scale=0.25]{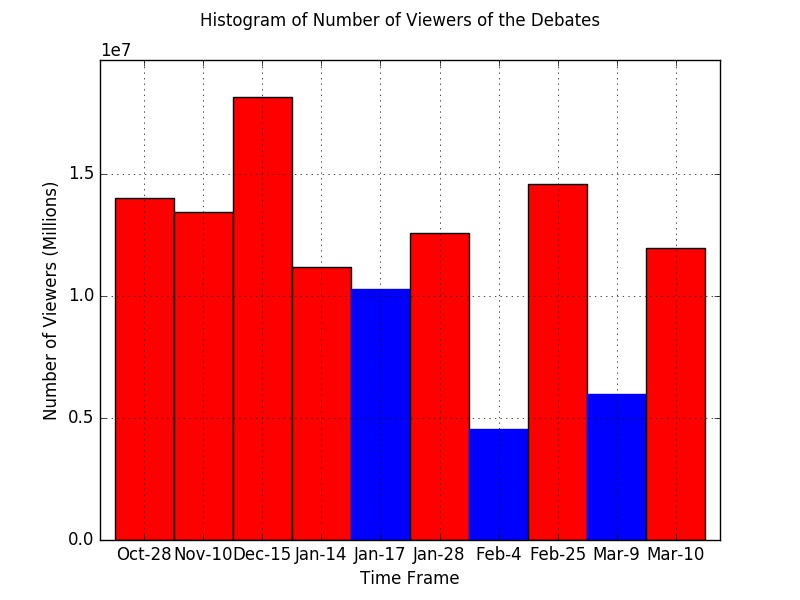}
\label{fig:viewer_debate}}
\caption{Histograms of Tweet Frequency vs. Debates and TV Viewers vs. Debates shown side-by-side for comparison. The \textbf{red} bars correspond to the Republican debates and the \textbf{blue} bars correspond to the Democratic debates.}
\label{fig:viewer-vs-tweets}
\end{figure*}
        
\begin{figure}[t]
\centering
\includegraphics[width=0.3\textwidth]{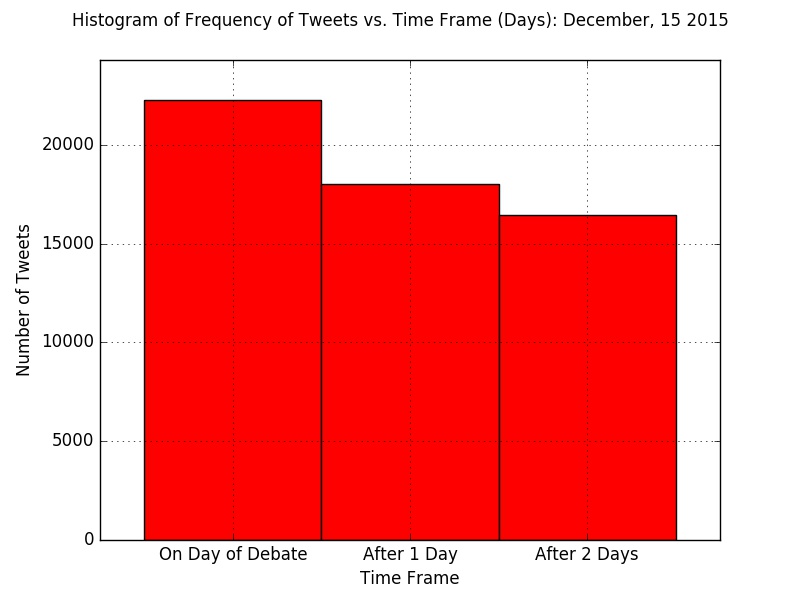}
\caption{Tweet Frequency vs. Days for the 5th Republican Debate (15th December 2015).}
\label{fig:tweet_days}
\end{figure}

\begin{figure}[t]
\centering
\includegraphics[width=0.3\textwidth]{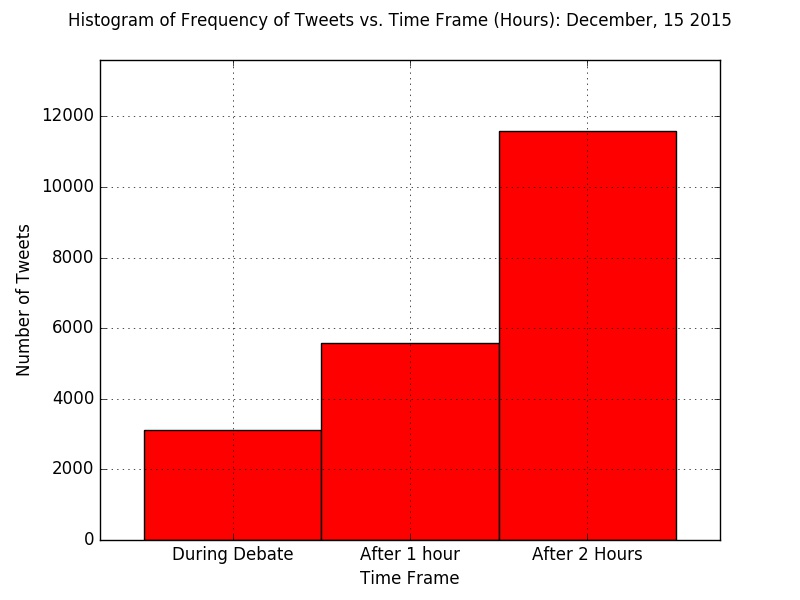}
\caption{Tweet Frequency vs. Hours for the 5th Republican Debate (15th December 2015).} 
\label{fig:tweet_hours}
\end{figure}

First, we look at the trend of the tweet frequency. Figure \ref{fig:viewer-vs-tweets} shows the trends of the tweet frequency and the number of TV viewers as the debates progress over time. We observe from Figures \ref{fig:tweet_debate} and \ref{fig:viewer_debate} that for the first $5$ debates considered, the trend of the number of TV viewers matches the trend of the number of tweets. However, we can see that towards the final debates, the frequency of the tweets decreases consistently. This shows an interesting fact that although the people still watch the debates, the number of people who tweet about it are greatly reduced. But the tweeting community are mainly youngsters and this shows that most of the tweeting community, who actively tweet, didn't watch the later debates. Because if they did, then the trends should ideally match.

Next we look at how the tweeting activity is on days of the debate: specifically on the day of the debate, the next day and two days later. Figure \ref{fig:tweet_days} shows the trend of the tweet frequency around the day of the 5th republican debate, i.e December 15, 2015. As can be seen, the average number of people tweet more on the day of the debate than a day or two after it. This makes sense intuitively because the debate would be fresh in their heads.

Then, we look at how people tweet in the hours of the debate: specifically during the debate, one hour after and then two hours after. Figure \ref{fig:tweet_hours} shows the trend of the tweet frequency around the hour of the 5th republican debate, i.e December 15, 2015. We notice that people don't tweet much during the debate but the activity drastically increases after two hours. This might be because people were busy watching the debate and then taking some time to process things, so that they can give their opinion. 

\begin{figure}[th]
\centering
\includegraphics[width=0.45\textwidth]{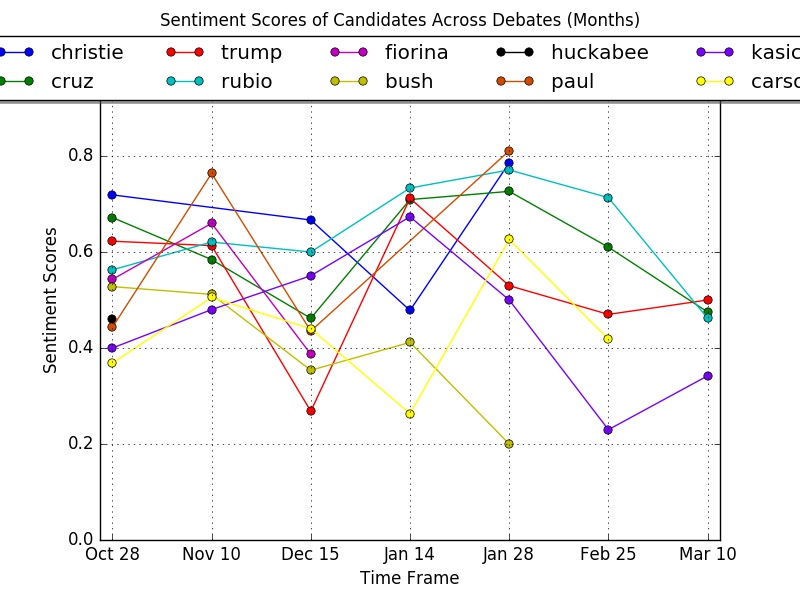}
\caption{Sentiments of the users towards the candidates across Debates.} 
\label{fig:sentiment_debate}
\end{figure}

\begin{figure*}[!t]
\centering
\subfloat[Sentiments of the users towards the candidates across Days. This is for the 5th Republican Debate: Dec 15, 2015]{\includegraphics[scale=0.27]{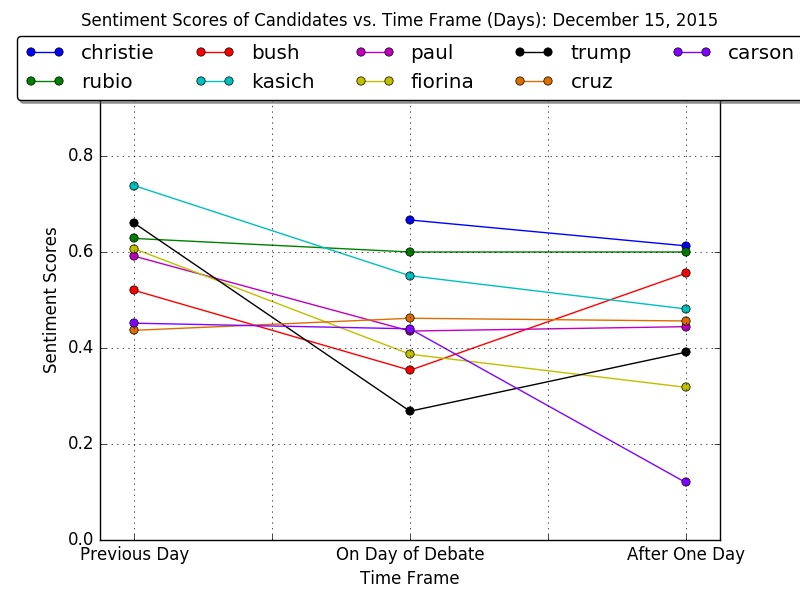}
\label{fig:sentiment_days_1}}
\hfil
\subfloat[Sentiments of the users towards the candidates across Days. This is for the 12th Republican Debate: March 10, 2016]{\includegraphics[scale=0.27]{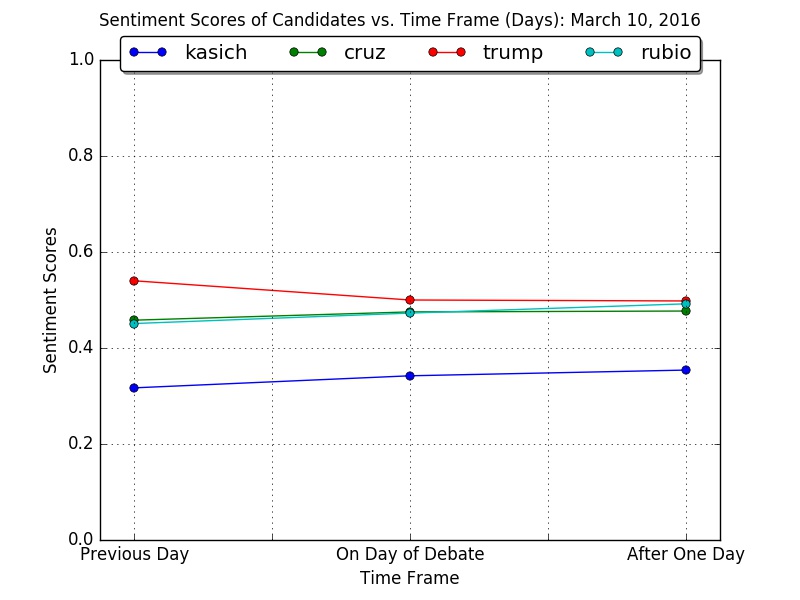}
\label{fig:sentiment_days_2}}
\caption{Graphs showing how the sentiments of the users towards the candidates before and after the debates.}
\label{fig:sentiment_days}
\end{figure*}

We have seen the frequency of tweets over time in the previous trends. Now, we will look at how the sentiments of the candidates change over time. 

First, Figure \ref{fig:sentiment_debate} shows how the sentiments of the candidates changed across the debates.\footnote{We mainly focus on the Republican Party here. Similar trends, although not shown here, are observed for the Democratic Party as well.} We find that many of the candidates have had ups and downs towards in the debates. But these trends are interesting in that, it gives some useful information about what went down in the debate that caused the sentiments to change (sometimes drastically). For example, if we look at the graph for Donald Trump, we see that his sentiment was at its lowest point during the debate held on December 15. Looking into the debate, we can easily see why this was the case. At a certain point in the debate,  Trump was asked about his ideas for the nuclear triad.\footnote{A nuclear triad refers to the nuclear weapons delivery of a strategic nuclear arsenal which consists of three components, traditionally strategic bombers, intercontinental ballistic missiles (ICBMs), and submarine-launched ballistic missiles (SLBMs).} It is very important that a presidential candidate knows about this, but Trump had no idea what the nuclear triad was and, in a transparent attempt to cover his tracks, resorted to a ``we need to be strong" speech. It can be due to this embarrassment that his sentiment went down during this debate.

Next, we investigate how the sentiments of the users towards the candidates change before and after the debate. In essence, we examine how the debate and the results of the debates given by the experts affects the sentiment of the candidates. Figure \ref{fig:sentiment_days_1} shows the sentiments of the users towards the candidate during the $5^{th}$ Republican Debate, 15th December 2015. It can be seen that the sentiments of the users towards the candidates does indeed change over the course of two days. One particular example is that of Jeb Bush. It seems that the populace are generally prejudiced towards the candidates, which is reflected in their sentiments of the candidates on the day of the debate. The results of the Washington Post are released in the morning after the debate. One can see the winners suggested by \textit{the Washington Post} in Table \ref{table:outcomes_debates}. One of the winners in that debate according to them is Jeb Bush. Coincidentally, Figure \ref{fig:sentiment_days_1} suggests that the sentiment of Bush has gone up one day after the debate (essentially, one day after the results given by the experts are out). 

There is some influence, for better or worse, of these experts on the minds of the users in the early debates, but towards the final debates the sentiments of the users are mostly unwavering, as can be seen in Figure \ref{fig:sentiment_days_2}. Figure \ref{fig:sentiment_days_2} is on the last Republican debate, and suggests that the opinions of the users do not change much with time. Essentially the users have seen enough debates to make up their own minds and their sentiments are not easily wavered.

\section{Evaluation Metrics}
\label{sec:metrics}
In this section, we define the different evaluation metrics that we use for different tasks. We have two tasks at hand: 1) Sentiment Analysis, 2) Outcome Prediction. We use different metrics for these two tasks.
\subsection{Sentiment Analysis}
In the study of sentiment analysis, we use ``Hamming Loss'' to evaluate the performance of the different methods. Hamming Loss, based on Hamming distance, takes into account the prediction error and the missing error, normalized over the total number of classes and total number of examples \cite{sorower2010literature}. The Hamming Loss is given below:

\begin{equation}
    Hamming\; Loss = \frac{1}{|D|} \sum\limits_{i=1}^{|D|} \frac{S_i \oplus Y_i}{|L|},
\end{equation}
where $|D|$ is the number of examples in the dataset and $|L|$ is the number of labels. $S_i$ and $Y_i$ denote the sets of true and predicted labels for instance $i$ respectively. $\oplus$ stands for the \texttt{XOR} operation \cite{tsoumakas2006multi}. Intuitively, the performance is better, when the Hamming Loss is smaller. $0$ would be the ideal case. 

\subsection{Outcome Prediction}
For the case of outcome prediction, we will have a predicted set and an actual set of results. Thus, we can use common information retrieval metrics to evaluate the prediction performance. Those metrics are listed below:

\textbf{Mean F-measure}: F-measure takes into account both the precision and recall of the results. In essence, it takes into account how many of the relevant results are returned and also how many of the returned results are relevant. 
	\begin{equation}
	    Mean \; F\text{-}measure = \frac{1}{|D|} \sum\limits_{i=1}^{|D|} 2 \frac{P_i \times R_i}{P_i + R_i},
	\end{equation}
where $|D|$ is the number of queries (debates/categories for grammy winners etc.), $P_i$ and $R_i$ are the precision and recall for the $i^{th}$ query.
	
\textbf{Mean Average Precision}: As a standard metric used in information retrieval, Mean Average Precision for a set of queries is mean of the average precision scores for each query:
	\begin{equation}
	    MAP = \frac{1}{|D|} \sum\limits_{i=1}^{|D|} \frac{\sum_{k=1}^n (P_i(k) \times \text{rel}_i(k))}{|RD_i|},
	\end{equation}
	where $|D|$ is the number of queries (e.g., debates), $P_i(k)$ is the precision at $k$ ($P@k$) for the $i^{th}$ query, $rel_i(k)$ is an indicator function, equaling $1$ if the document at position $k$ for the $i^th$ query is relevant, else $0$, and $|RD_i|$ is the number of relevant documents for the $i^{th}$ query.

\section{Results}
\label{sec:results}
\subsection{Sentiment Analysis}
We use $3$ different datasets for the problem of sentiment analysis, as already mentioned. We test the four different algorithms mentioned in Section \ref{subsec:ml_methods}, with a different combination of features that are described in Section \ref{subsec:features}. To evaluate our models, we use the ``Hamming Loss'' metric as discussed in Section~\ref{sec:metrics}. We use this metric because our problem is in the multi-class classification domain. However, the single-label classifiers like SVM, Naive Bayes, Elman RNN cannot be evaluated against this metric directly. To do this, we split the predicted labels into a format that is consistent with that of multi-label classifiers like RaKel. The results of the experiments for each of the datasets are given in Tables \ref{table:sentiments}, \ref{table:sentiments_grammy} and \ref{table:sentiments_superbowl}. In the table, $f_1$, $f_2$, $f_3$, $f_4$, $f_5$ and $f_6$ denote the features $1$-$gram$, $2$-$gram$, $(1+2)$-$gram$, $(1+2)$-$gram + MISC$, $DOC$ and $DOC + TOPIC$ respectively. Note that lower values of hamming losses are more desirable.

We find that RaKel performs the best out of all the algorithms. RaKel is more suited for the task because our task is a multi-class classification. Among all the single-label classifiers, SVM performs the best. 
We also observe that as we use more complex feature spaces, the performance increases. This is true for almost all of the algorithms listed. 
    
Our best performing features is a combination of paragraph embedding features and topic features from LDA. This makes sense intuitively because paragraph-embedding takes into account the context in the text. Context is very important in determining the sentiment of a short text: having negative words in the text does not always mean that the text contains a negative sentiment. For example, the sentence ``never say never is not a bad thing'' has many negative words; but the sentence as a whole does not have a negative sentiment. This is why we need some kind of context information to accurately determine the sentiment. Thus, with these embedded features, one would be able to better determine the polarity of the sentence. The other label is the entity (candidate/song etc.) in consideration. Topic features here make sense because this can be considered as the topic of the tweet in some sense. This can be done because that label captures what or whom the tweet is about.

\begin{table}[ht]
\vspace{2ex}
\setlength\tabcolsep{4pt}
\begin{minipage}{0.48\textwidth}
\begin{tabular}{l | l | l | l | l | l | l}
\toprule
& \multicolumn{6}{c}{\textbf{Hamming Loss}}\\
\textbf{Method} & $f_1$ & $f_2$ & $f_3$ & $f_4$ & $f_5$ & $f_6$\\
\midrule
Naive Bayes & 0.521 & 0.523 & 0.518 & 0.520 & 0.487 & 0.489\\
SVM & 0.488 & 0.489 & 0.483 & 0.482 & 0.439 & 0.437\\
Elman RNN & 0.491 & 0.490 & 0.484 & 0.485 & 0.446 & 0.442\\
RaKel & 0.455 & 0.454 & 0.449 & 0.443 & 0.394 & 0.393\\
\bottomrule
\end{tabular}
\centering
\caption{Sentiment Analysis for the Presidential Debates: $f_1$ stands for $1$-$gram$, $f_2$ stands for $2$-$gram$, $f_3$ stands for $(1+2)$-$gram$, $f_4$ stands for $(1+2)$-$gram + MISC$, $f_5$ stands for $DOC$, $f_6$ stands for $DOC + TOPIC$.}
\label{table:sentiments}

\end{minipage}%
\hfill
\begin{minipage}{0.48\textwidth}
\begin{tabular}{l | l | l | l | l | l | l}
\toprule
& \multicolumn{6}{c}{\textbf{Hamming Loss}}\\
\textbf{Method} & $f_1$ & $f_2$ & $f_3$ & $f_4$ & $f_5$ & $f_6$\\
\midrule
Naive Bayes & 0.542 & 0.554 & 0.545 & 0.541 & 0.482 & 0.480\\
SVM & 0.489 & 0.494 & 0.481 & 0.492 & 0.446 & 0.447\\
Elman RNN & 0.493 & 0.502 & 0.491 & 0.489 & 0.451 & 0.450\\
RaKel & 0.468 & 0.466 & 0.461 & 0.457 & 0.398 & 0.396\\
\bottomrule
\end{tabular}
\centering
\caption{Sentiment Analysis for the 2013 Grammy Awards}
\label{table:sentiments_grammy}
\end{minipage}
\hfill
\begin{minipage}{0.48\textwidth}
\begin{tabular}{l | l | l | l | l | l | l}
\toprule
& \multicolumn{6}{c}{\textbf{Hamming Loss}}\\
\textbf{Method} & $f_1$ & $f_2$ & $f_3$ & $f_4$ & $f_5$ & $f_6$\\
\midrule
Naive Bayes & 0.504 & 0.512 & 0.507 & 0.506 & 0.461 & 0.465\\
SVM & 0.467 & 0.469 & 0.462 & 0.460 & 0.433 & 0.431\\
Elman RNN & 0.481 & 0.483 & 0.475 & 0.472 & 0.436 & 0.437\\
RaKel & 0.465 & 0.462 & 0.459 & 0.439 & 0.385 & 0.383\\
\bottomrule
\end{tabular}
\centering
\caption{Sentiment Analysis for the 2013 Superbowl}
\label{table:sentiments_superbowl}
\end{minipage}

\end{table}

\begin{table*}[ht]
\vspace{2ex}
\resizebox{\textwidth}{!}{\begin{tabular}{l | l | l | l}
\toprule
\textbf{Debate} & \textbf{Candidates Predicted} & \textbf{Winners by Washington Post} & \textbf{Participants}\\
\midrule
Oct 28-R & $\text{Rubio}^4$, $\text{Cruz}^2$, $\text{Christie}^1$, $\text{Trump}^3$ & Rubio, Cruz, Christie, Trump & Trump, Carson, Rubio, Bush, Fiorina, Cruz, Huckabee, Christie, Kasich, Paul\\
Nov 10-R & $\text{Rubio}^3$, $\text{Cruz}^5$, $\text{Trump}^4$, $\text{Fiorina}^2$, $\text{Paul}^1$ & Rubio, Cruz, Carson, Fiorina, Paul & Trump, Carson, Rubio, Bush, Fiorina, Cruz, Kasich, Paul\\
Dec 15-R & $\text{Cruz}^4$, $\text{Rubio}^2$, $\text{Christie}^1$, $\text{Kasich}^3$,  & Bush, Rubio, Christie, Trump (1st Hour) & Trump, Carson, Rubio, Bush, Fiorina, Cruz, Kasich, Paul, Christie\\
Jan 14-R & $\text{Rubio}^1$, $\text{Trump}^2$, $\text{Cruz}^3$ & Rubio, Trump, Cruz & Trump, Carson, Rubio, Bush, Cruz, Kasich, Christie\\
Jan 17-D & $\text{Sanders}^1$, $\text{Malley}^2$ & Sanders, Malley & Hillary, Sanders, Malley\\
Jan 28-R & $\text{Paul}^1$, $\text{Christie}^2$ & Paul, Bush & Trump, Carson, Rubio, Bush, Cruz, Kasich, Christie, Paul\\
Feb 4-D & $\text{Sanders}^1$ & Sanders & Hillary, Sanders\\
Feb 25-R & $\text{Rubio}^1$ & Rubio & Trump, Carson, Rubio, Cruz, Kasich\\
March 9-D & $\text{Sanders}^1$ & Hillary & Hillary, Sanders\\
March 10-R & $\text{Rubio}^3$, $\text{Trump}^1$, $\text{Cruz}^2$ & Rubio, Trump, Cruz & Trump, Rubio, Cruz, Kasich\\
\bottomrule
\end{tabular}}
\centering
\caption{Outcome Prediction based on Tweet Sentiment: the superscript on the candidates indicates the predicted ordering}
\label{table:outcomes_debates}
\end{table*}

\begin{table*}[ht]
\vspace{2ex}
\resizebox{\textwidth}{!}{\begin{tabular}{l | l | l | l}
\toprule
\textbf{Category} & \textbf{Predicted Winner} & \textbf{Actual Winner} & \textbf{Nominees}\\
\midrule
Album of the Year & Babel & Babel & Some Nights, Blunderbuss, El Camino, Babel, Channel Orange\\\hline
Song of the Year & Call Me Maybe & We Are Young & Adorn, Call Me Maybe, The A Team, We Are Young, Stronger\\\hline
Best New Artist & Fun & Fun & Frank Ocean, The Lumineers, Hunter Hayes, Fun, Alabama Shakes\\\hline
Record of the Year & Somebody That I Used to Know & Somebody That I Used to Know & Lonely Boy, Somebody That I Used to Know, We are Never Ever Getting\\
& & & Back Together, Thinkin Bout You, We Are Young, Stronger\\
\bottomrule
\end{tabular}}
\centering
\caption{Outcome Prediction for the 2013 Grammy awards.}
\label{table:outcomes_grammy}
\end{table*}

\begin{table}[ht]
\vspace{2ex}
\resizebox{0.5\textwidth}{!}
{\begin{tabular}{l | l | l | l}
\toprule
\textbf{Category} & \textbf{Predicted Winner} & \textbf{Actual Winner} & \textbf{Participants}\\
\midrule
Overall Winner & Baltimore Ravens & Baltimore Ravens & Baltimore Ravens, San Francisco 49ers\\
Most Valuable Player & Joe Flacco & Joe Flacco & Players from both the teams\\
\bottomrule
\end{tabular}}
\centering
\caption{Outcome Prediction for Superbowl 2013}
\label{table:outcomes_superbowl}
\end{table}

\subsection{Results for Outcome Prediction}
In this section, we show the results for the outcome-prediction of the events. RaKel, as the best performing method, is trained to predict the sentiment-labels for the unlabeled data. The sentiment labels are then used to determine the outcome of the events. In the Tables (\ref{table:outcomes_debates}, \ref{table:outcomes_grammy}, \ref{table:outcomes_superbowl}) of outputs given, we only show as many predictions as there are winners.

\subsubsection{Presidential Debates}
The results obtained for the outcome prediction task for the US presidential debates is shown in Table \ref{table:outcomes_debates}. Table \ref{table:outcomes_debates} shows the winners as given in the Washington Post (3rd column) and the winners that are predicted by our system (2nd column).\footnote{The winners given by the Washington Post are not ranked, as opposed to our model.} By comparing the \textit{set} of results obtained from both the sources, we find that the set of candidates predicted match to a large extent with the winners given out by the Washington Post. 
The result suggests that the opinions of the social media community match with that of the journalists in most cases.

\subsubsection{Grammy Awards}
A Grammy Award is given to outstanding achievement in the music industry. There are two types of awards: ``General Field'' awards, which are not restricted by genre, and genre-specific awards. Since, there can be upto $80$ categories of awards, we only focus on the main $4$: 1) Album of the Year, 2) Record of the Year, 3) Song of the Year, and 4) Best New Artist. These categories are the main in the awards ceremony and most looked forward to. That is also why we choose to predict the outcomes of these categories based on the tweets. We use the tweets before the ceremony (but after the nominations) to predict the outcomes.

Basically, we have a list of nominations for each category. We filter the tweets based on these nominations and then predict the winner as with the case of the debates. The outcomes are listed in Table \ref{table:outcomes_grammy}. We see that largely, the opinion of the users on the social network, agree with the deciding committee of the awards. The winners agree for all the categories except ``Song of the Year''.

\subsubsection{Super Bowl}
The Super Bowl is the annual championship game of the National Football League. We have collected the data for the year 2013. Here, the match was between the Baltimore Ravens and the San Francisco 49ers. The tweets that we have collected are during the game. From these tweets, one could trivially determine the winner. But an interesting outcome would be to predict the \textit{Most Valuable Player} (MVP) during the game. To determine this, all the tweets were looked at and we determined the candidate with the highest positive sentiment by the end of the game. The result in Table \ref{table:outcomes_superbowl} suggests that we are able to determine the outcomes accurately.

\begin{table}
\vspace{2ex}
\begin{tabular}{l | c | c |c}
\toprule
\textbf{Metric} & \textbf{Debates} & \textbf{Grammys} & \textbf{Super Bowl}\\
\midrule
Mean F1-Measure & 0.636 & 0.750 & 1.000\\
Mean Average Precision & 0.677 & 0.458 & 0.590\\
\bottomrule
\end{tabular}
\centering
\caption{Metric Results for Outcome Prediction}
\label{table:outcomes_metrics}
\end{table}

Table \ref{table:outcomes_metrics} displays some evaluation metrics for this task. These were computed based on the predicted outcomes and the actual outcomes for each of the different datasets. Since the number of participants varies from debate-to-debate or category-to-category for Grammy etc., we cannot return a fixed number of winners for everything. So, the size of our returned ranked-list is set to half of the number of participants (except for the MVP for Super Bowl; there are so many players and returning half of them when only one of them is relevant is meaningless. So, we just return the top $10$ players). As we can see from the metrics, the predicted outcomes match quite well with the actual ones (or the ones given by the experts). 



\section{Conclusions}
\label{sec:conclusions}
This paper presents a study that compares the opinions of users on microblogs, which is essentially the crowd wisdom, to that of the experts in the field. Specifically, we explore three datasets: US Presidential Debates 2015-16, Grammy Awards 2013, Super Bowl 2013. We determined if the opinions of the crowd and the experts match by using the sentiments of the tweets to predict the outcomes of the debates/Grammys/Super Bowl. 
We observed that in most of the cases, the predictions were right indicating that crowd wisdom is indeed worth looking at and mining sentiments in microblogs is useful. In some cases where there were disagreements, however, we observed that the opinions of the experts did have some influence on the opinions of the users. We also find that the features that were most useful in our case of multi-label classification was a combination of the document-embedding and topic features. 

\bibliography{acl2016.bib}
\bibliographystyle{IEEEtran.bst}


\end{document}